\documentclass{article}

\usepackage{microtype}
\usepackage{graphicx}
\usepackage{subcaption}
\usepackage{booktabs}
\usepackage{hyperref}

\usepackage[accepted]{icml2026}
\makeatletter
\renewcommand{\ICML@appearing}{Mechanistic Interpretability Workshop at the $\mathit{43}^{rd}$ International Conference on Machine Learning, Seoul, South Korea, 2026. Copyright 2026 by the author(s).}
\makeatother
\usepackage{amsmath}
\usepackage{amssymb}
\usepackage{multirow}
\usepackage[capitalize,noabbrev]{cleveref}

\icmltitlerunning{Geometry of Ordinal Representations in LMs}

\begin{document}

\twocolumn[
\icmltitle{Geometry of Ordinal Representations in Language Models}

\icmlsetsymbol{equal}{*}

\begin{icmlauthorlist}
\icmlauthor{Saksham Bassi}{equal}
\icmlauthor{Sharvi Tomar}{equal}
\icmlcorrespondingauthor{Sharvi Tomar}{stomar2@illinois.edu}
\end{icmlauthorlist}

\vskip 0.3in
]

\printAffiliationsAndNotice{\icmlEqualContribution}

\begin{abstract}
Recent work showed that language models represent character counts on curved 1D manifolds, with attention heads performing geometric transformations to enable computation. We test whether this generalizes across four ordinal tasks (bracket depth, indentation, table position, numeric magnitude) in Gemma-2-2B, Gemma-2-9B, and Qwen3-4B. We find that 1D manifolds with place-cell feature tiling emerge for tasks where the ordinal variable is locally computable from token identity, while tasks requiring cross-position integration or semantic extraction produce higher-dimensional or incoherent representations. Geometric computation is architecture-dependent: Qwen3-4B shows substantially stronger twisting than Gemma models for indentation, and its twisters preserve ordinal order, unlike its numeric twisters. Activation patching confirms that the identified manifold subspaces concentrate task-relevant information, with manifold-direction ablation causing dramatically larger probe accuracy drops than random-direction controls.
\end{abstract}

%% ─── INTRODUCTION ───
\section{Introduction}
\label{sec:intro}

Understanding how neural networks internally represent structured information is a central goal of mechanistic interpretability \citep{olah2020zoom, conmy2023automated}. While much work has focused on identifying circuits and features \citep{elhage2022superposition, bricken2023monosemanticity, templeton2024scaling}, recent studies have begun characterizing the \emph{geometry} of learned representations.

\citet{gurnee2026manifolds} demonstrated that Claude 3.5 Haiku represents character counts on curved 1D manifolds in the residual stream, discretized by sparse autoencoder (SAE) features that tile the manifold like place cells tile physical space \citep{moser2008place}. Attention heads perform geometric transformations (twisting and aligning the manifold) to compute distance to line boundaries. This raises a natural question: do these geometric motifs generalize to other tasks requiring ordinal or counting representations, and do different model architectures learn the same geometry?

We investigate four ordinal tasks across three open-weight models: Gemma-2-2B \citep{team2024gemma}, Gemma-2-9B \citep{team2024gemma}, and Qwen3-4B \citep{qwen2025qwen3}. The tasks are bracket nesting depth, Python indentation level, markdown table position (row and column), and numeric magnitude. For each task and model, we (1)~train linear probes to verify decodability, (2)~discover manifold subspaces via PCA on per-value mean activations, (3)~find SAE feature families using Gemma Scope \citep{lieberum2024gemma} for the Gemma models, (4)~analyze attention heads for geometric computation (twisting and alignment), and (5)~test the necessity of manifold subspaces via activation patching.

Our main findings:
\begin{itemize}
    \item \textbf{Manifold hypothesis partially generalizes:} Bracket depth and table column index produce 1D manifolds with place-cell feature tiling across all models. Indentation uses a 3D subspace. Numeric magnitude shows no coherent manifold despite strong linear decodability.
    \item \textbf{Geometric computation is architecture-dependent:} Bracket depth manifolds are pre-organized from the embedding layer in all three models. Indentation and numeric magnitude show dramatically more twisting in Qwen3-4B than in either Gemma model, despite similar probing accuracy.
    \item \textbf{Coherent vs.\ incoherent twisting:} High twist alone does not imply ordinal geometric computation. Qwen3-4B's indentation twisters preserve ordinal order, while its numeric twisters do not ($\rho \leq 0.23$). This distinguishes heads that geometrically construct an ordinal manifold from heads that merely rearrange representations without structure.
    \item \textbf{Subspace ablation confirms informational concentration:} Activation patching of the top-$k$ manifold directions causes dramatically larger probe accuracy drops than random-direction controls across all tasks and all three models, establishing that the identified subspaces are necessary for linear decodability of the ordinal variable.
    \item \textbf{A key factor is local computability.} Tasks where the ordinal variable is directly computable from token identity (brackets, table columns) produce clean 1D manifolds that require no geometric construction. Tasks requiring cross-position integration or semantic extraction produce higher-dimensional or incoherent representations.
\end{itemize}

%% ─── RELATED WORK ───
\section{Related Work}
\label{sec:related}

\paragraph{Representation geometry.}
The study of geometric structure in neural network representations has a rich history. \citet{hewitt2019structural} introduced structural probing to find syntax in word representations, establishing the methodology of projecting representations into task-relevant subspaces. The logit lens \citep{nostalgebraist2020logitlens} and tuned lens \citep{belrose2023tunedlens} revealed that intermediate representations progressively converge toward output predictions. \citet{marks2023geometry} found emergent linear structure in LLM representations of truth values. \citet{gurnee2024spacetime} showed that LLMs represent continuous variables (space and time) as linear directions. \citet{park2023linear} formalized the linear representation hypothesis, and \citet{engels2025notall} extended it, showing that some features require multi-dimensional subspaces rather than single directions. \citet{henighan2025manifolds} argued that features are represented as manifolds and that cosine similarity encodes intrinsic manifold geometry. \citet{vafa2024belief} showed that transformers represent belief state geometry in their residual stream, including non-trivial fractal structures. \citet{gurnee2026manifolds} made this concrete for counting tasks, showing that character counts trace curved 1D manifolds with SAE features acting as place-cell analogs. Their analysis of attention heads operates in Q-space, i.e. the subspace obtained by projecting residual stream activations through each head's query weight matrix, which we adopt in Section \ref{sec:methods}. Our work directly extends this to multiple tasks and multiple open-weight models.

\paragraph{Theoretical grounding.}
\citet{shai2026factored} proved that transformers learn factored representations in orthogonal subspaces, with dimensionality growing linearly when factors are conditionally independent. This predicts the orthogonal arrangement observed by \citet{gurnee2026manifolds} and suggests it should generalize across tasks. \citet{park2026infogeo} argued that representation geometry should be understood information-geometrically rather than in Euclidean terms, which is relevant for interpreting our curved manifold findings and for the causal steering experiments we perform.

\paragraph{Sparse autoencoders and features.}
Sparse autoencoders have emerged as a primary tool for decomposing neural network representations into interpretable features \citep{cunningham2023sparse, bricken2023monosemanticity}. \citet{templeton2024scaling} scaled this approach to Claude 3 Sonnet, finding interpretable features at scale. Gemma Scope \citep{lieberum2024gemma} provides pretrained SAEs for Gemma-2 models, which we use for feature family discovery. \citet{michaud2025saescaling} studied how SAE scaling interacts with feature manifolds, finding that manifold structure affects the number of features needed; our results on numeric magnitude (high selectivity but zero monotonic features) are consistent with their finding that SAEs struggle to decompose manifold-structured features into clean families. \citet{lubana2025priors} noted that SAEs assume temporal independence, potentially missing sequential structure in how manifolds evolve across positions, which is a relevant caveat for our ordinal tasks.

\paragraph{Mechanistic interpretability of structured tasks.}
Prior work has studied how transformers represent structured information including modular arithmetic \citep{nanda2023progress} and algorithmic tasks \citep{conmy2023automated}. \citet{beckmann2025understanding} proposed a tiered framework for mechanistic understanding; if multiple tasks and multiple architectures use the same geometric motifs, this would constitute evidence for principled rather than task-specific or architecture-specific representations. Our cross-model comparison directly addresses this question.

\paragraph{Cross-architecture comparison.}
Gemma-2 \citep{team2024gemma} uses grouped-query attention, SwiGLU activations, and alternating local/global attention. Qwen3-4B \citep{qwen2025qwen3} uses grouped-query attention, RoPE positional encoding, and SwiGLU activations, with a substantially larger head count (32 heads vs.\ 8 in Gemma-2-2B). These architectural differences, particularly the head count and positional encoding scheme, may explain the differences in geometric computation we observe.

%% ─── METHODS ───
\section{Methods}
\label{sec:methods}

\subsection{Models and Tasks}

We study three open-weight models: Gemma-2-2B (2.6B parameters, 26 layers, 8 heads), Gemma-2-9B (9.2B parameters, 42 layers, 16 heads), and Qwen3-4B (4.0B parameters, 36 layers, 32 heads). Qwen3-4B uses grouped-query attention and RoPE positional encoding. We use Gemma Scope \citep{lieberum2024gemma} residual stream SAEs (16K dictionary) for the Gemma models; no pretrained SAEs exist for Qwen3-4B.

We study four ordinal tasks. \textbf{Bracket nesting depth}: sequences of mixed brackets with per-token depth labels (0--10); 3,000 samples, half synthetic and half from The Stack. \textbf{Python indentation level}: per-token indent level (0--5 in units of 4 spaces); 3,000 samples from The Stack. \textbf{Markdown table position}: synthetic tables with per-token row (0--14) and column (0--8) indices; 2,000 tables. \textbf{Numeric magnitude}: text with numbers labeled by $\lfloor\log_{10}(\text{value})\rfloor$ (0--8); 5,000 samples from OpenWebText. All sequences are tokenized with each model's native tokenizer (max length 256).

\subsection{Probing, Manifold Discovery, and Feature Families}

For each task, model, and layer $\ell$, we collect residual stream activations at labeled token positions (4,000 train / 1,000 test). We train logistic regression probes (classification) or Ridge regression probes (numeric magnitude) and report test accuracy or $R^2$.

For manifold discovery, we compute the mean activation $\bar{\mathbf{h}}_v = \frac{1}{|S_v|}\sum_{i \in S_v} \mathbf{h}_i$ for each ordinal value $v$ and apply PCA to the matrix of per-value means. We report the number of Principal Components (PCs) needed to explain 90\% of variance and pairwise cosine similarities between per-value means. We note that this characterizes the geometry of \emph{class centroids}: high PC1 variance among centroids is necessary but not sufficient for individual tokens to concentrate near a 1D manifold, as it is consistent with heterogeneous or multimodal within-class distributions. We use this centroid geometry because it captures how the model \emph{distinguishes} ordinal values; the subspace ablation results (Section \ref{sec:results}) provide complementary evidence that these directions are informationally concentrated.

For the Gemma models, we use Gemma Scope SAEs at four evenly spaced layers to find feature families. For each feature $j$, the mean activation per ordinal value gives a tuning curve $\tau_j(v)$. We rank features by between-value variance, keeping those active on $>5\%$ of tokens. A feature is \emph{selective} if its peak exceeds $3\times$ its mean, and \emph{monotonic} if its tuning curve is approximately non-decreasing or non-increasing.

\subsection{Geometric Computation and Subspace Ablation}

Following \citet{gurnee2026manifolds}, we measure how much each attention head rearranges the manifold geometry by projecting the per-value activation means into the query (Q) space. Formally, for an attention head $h$ in layer $\ell$ with query weight matrix $\mathbf{W}_Q^h \in \mathbb{R}^{d_{\text{query}} \times d_{\text{model}}}$, the projected mean for ordinal value $v$ is computed as:
\begin{equation}
\bar{\mathbf{q}}_v^h = \mathbf{W}_Q^h \bar{\mathbf{h}}_v
\end{equation}
where $\bar{\mathbf{h}}_v \in \mathbb{R}^{d_{\text{model}}}$ is the mean residual stream activation vector for value $v$.

Using these projections, we compute three distinct diagnostic scores for each attention head:
\begin{itemize}
    \item \textbf{Twist Score}: Defined as $1 - \text{corr}(\mathbf{d}_{\text{resid}}, \mathbf{d}_Q)$, where $\mathbf{d}_{\text{resid}}$ is the vector of all pairwise Euclidean distances between the per-value means in the original residual stream ($\|\bar{\mathbf{h}}_u - \bar{\mathbf{h}}_v\|_2$ for all pairs $u, v$), $\mathbf{d}_Q$ is the vector of corresponding pairwise distances in the projected Q-space ($\|\bar{\mathbf{q}}_u^h - \bar{\mathbf{q}}_v^h\|_2$), and $\text{corr}$ denotes the Pearson correlation coefficient. A high twist score indicates that the attention head's projection significantly rearranges or distorts the relative geometric distances of the original manifold.
    
    \item \textbf{Alignment Score}: Defined as the fraction of total variance explained by the first principal component ($\text{PC1}$) of the Q-projected manifold centroids $\{\bar{\mathbf{q}}_v^h\}_v$. This measures the extent to which the head compresses the projected manifold into a dominant one-dimensional axis.
    
    \item \textbf{Ordinality Score}: Defined as $|\rho_{\text{Spearman}}(v, \text{PC1}(v))|$, where $\rho_{\text{Spearman}}$ is the Spearman rank correlation coefficient calculated between the true, ordered ordinal labels $v$ and the coordinates of their corresponding projected means along the first principal component $\text{PC1}$ of the Q-space. A high ordinality score indicates that the head preserves the strict monotonic progression of the ordinal sequence along its primary axis.
\end{itemize}

To test whether the manifold subspaces are necessary for linear decodability, we patch the top-$k$ PCA directions of the per-value means by replacing a token's projection onto this subspace with that of a token with a different ordinal label, leaving the orthogonal complement unchanged. We compare the resulting probe accuracy drop against patching $k$ random directions as a control. We report results at $k=3$ for the layer with the largest manifold drop. We follow best practices from \citet{heimersheim2024patching} and note the caveat of \citet{makelov2023subspace} that subspace patching can activate parallel pathways; our random-direction control is designed to mitigate this concern. We note that this methodology tests informational concentration in the subspace, not behavioral causality (i.e., whether the model's downstream generation depends on this subspace).

%% ─── RESULTS ───
\section{Results}
\label{sec:results}

\subsection{Manifold Structure and Feature Tiling}

Linear probes achieve high test accuracy for all tasks across all three models (\Cref{tab:probing}). Indentation and table columns reach more than $94\%$ test accuracy by mid-network in every model. Bracket depth peaks at 53--57\% test accuracy, which is low in absolute terms but well above chance (1/11 $\approx$ 9\%), and with near-perfect \emph{training} accuracy ($\geq 99\%$), indicating substantial overfitting. This gap likely arises because bracket-depth labels are highly correlated with token identity (specific bracket characters), creating spurious shortcuts the probe exploits on training tokens but that do not generalize to held-out tokens with different token-depth correlations. The manifold structure findings for brackets should be interpreted with this caveat: the subspace cleanly separates depth values in the training distribution, but a probe trained on it does not robustly generalize. Table rows reach 74--82\% test accuracy. Numeric magnitude achieves train $R^2 > 0.93$ in all models but test $R^2$ varies substantially: 0.54 for Gemma-2-2B, 0.55 for Gemma-2-9B at layer 0 (degrading to negative at deeper layers due to overfitting), and 0.69 for Qwen3-4B. The ordinal variable is linearly accessible in the residual stream for every task and model studied.

\begin{table}[t]
\caption{Best test probing score (layer in parentheses) across models. Metric is test accuracy except numeric (test $R^2$). 80/20 train/test split. Manifold dimensionality is shown in \Cref{fig:pc1_variance}.}
\label{tab:probing}
\centering
\small
\begin{tabular}{@{}lccc@{}}
\toprule
Task & Gemma-2-2B & Gemma-2-9B & Qwen3-4B \\
\midrule
Brackets    & 0.538 (L2)  & 0.525 (L6)  & 0.573 (L2)  \\
Table cols  & 0.983 (L12) & 0.996 (L24) & 0.992 (L16) \\
Indentation & 0.943 (L12) & 0.947 (L18) & 0.944 (L24) \\
Table rows  & 0.746 (L16) & 0.821 (L22) & 0.803 (L22) \\
Numeric     & 0.536 (L16) & 0.549 (L0)  & 0.688 (L2)  \\
\bottomrule
\end{tabular}
\end{table}

\begin{figure*}[t]
\centering
\includegraphics[width=\textwidth]{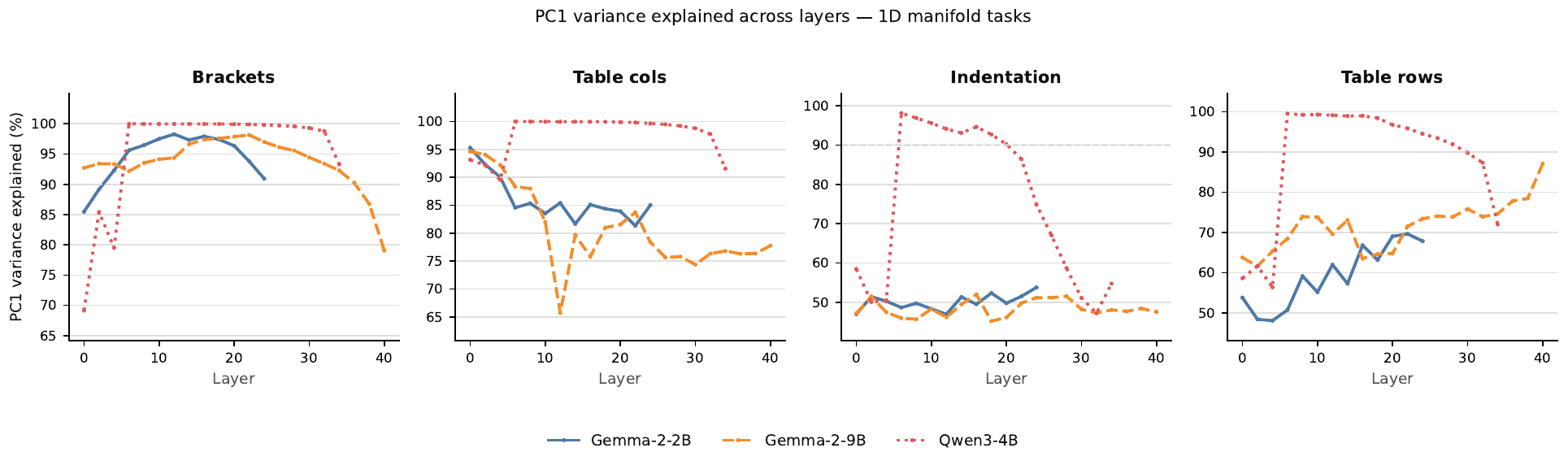}
\caption{\textbf{PC1 variance explained across layers.} For brackets and table columns, PC1 captures $>90\%$ of variance (dashed line) in all three models, consistent with 1D centroid manifold structure. For indentation and table rows, Qwen3-4B collapses to a 1D manifold at mid-layers (step-jump to $\sim$100\%) while Gemma models remain at 45--65\%, revealing architecture-dependent manifold organization.}
\label{fig:pc1_variance}
\end{figure*}

\paragraph{Manifold dimensionality.} The tasks separate into consistent tiers across all three models (\Cref{fig:pc1_variance}). Bracket depth and table column index concentrate more than $90\%$ of variance in PC1 at mid-to-late layers, replicating the curved 1D manifold found by \citet{gurnee2026manifolds} for character counts. Indentation requires 2--3 PCs for 90\% variance (PC1 $\approx$ 47--54\%), likely because indent level interacts with code structure. Table rows require 1--4 PCs depending on model, with Qwen3-4B collapsing to 1D at mid-layers while Gemma models use 3--4D. Numeric magnitude requires 4--5 PCs across all models (PC1 only 40--62\%), with no clear ordinal progression in the PCA projection. The existence of multi-dimensional manifolds is consistent with \citet{engels2025notall}, who showed that some LLM features require multi-dimensional subspaces.

\begin{figure*}[t]
\centering
\includegraphics[width=\textwidth]{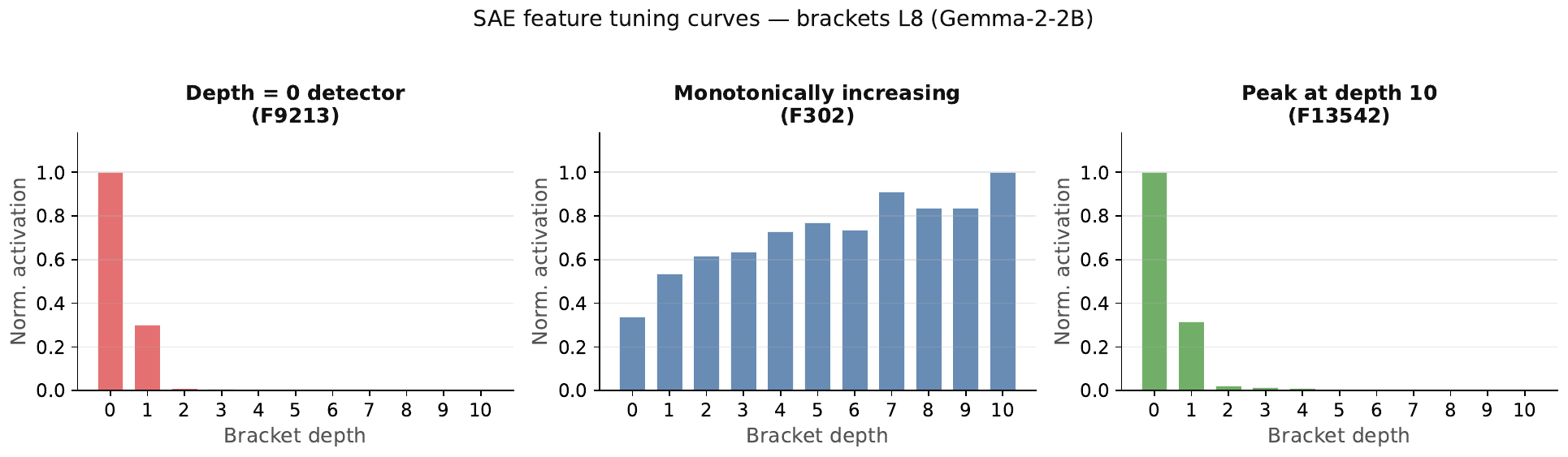}
\caption{\textbf{SAE feature tuning curves - brackets L8 (Gemma-2-2B).} Three archetypes of place-cell tiling: a sharp depth$=$0 detector (F9213), a monotonically increasing feature (F302), and a feature peaking at maximum depth (F13542). Activations normalized to $[0,1]$ per feature.}
\label{fig:tuning_curves}
\end{figure*}

\paragraph{Feature families (Gemma models).} SAE features tile the bracket-depth manifold cleanly (\Cref{tab:features}, \Cref{fig:tuning_curves}): at layer 8 in Gemma-2-2B, more than two-thirds of the top features are selective and include a sharp depth$=$0 detector, a monotonically increasing feature, and features peaking at specific depths. In Gemma-2-9B, roughly a third to half of the top features show similar place-cell structure. For indentation and table rows, selectivity (the fraction of top features with a peak exceeding three times their mean activation) is weaker (2–10 of the top 20) and features preferentially encode boundary values (indent$=$0, row$=$0) over fine-grained position. Numeric magnitude is the clear outlier: despite 6--12 out of 20 features being selective, \emph{zero} are monotonic at any layer in either model. Tuning curves are erratic, and the SAE cannot decompose numeric magnitude into an orderly feature family. This contrasts with \citet{heinzerling2024monotonic}, who found monotonic representations in encoder models, and is consistent with \citet{dehaene2003weber}'s observation that log-scale compression makes fine ordinal distinctions harder to encode discretely.

\begin{table}[b]
\caption{SAE feature family quality: selective and monotonic features out of top 20, at the layer with highest selectivity. Gemma models only (no pretrained SAEs for Qwen3-4B).}
\label{tab:features}
\centering
\small
\begin{tabular}{@{}lcccc@{}}
\toprule
 & \multicolumn{2}{c}{Gemma-2-2B} & \multicolumn{2}{c}{Gemma-2-9B} \\
\cmidrule(lr){2-3}\cmidrule(lr){4-5}
Task & Sel. & Mono. & Sel. & Mono. \\
\midrule
Brackets    & 15/20 (L14) & 4/20 & 9/20 (L4)  & 5/20 \\
Indentation & 7/20 (L8)   & 4/20 & 8/20 (L14) & 4/20 \\
Table rows  & 8/20 (L8)   & 2/20 & 10/20 (L4) & 1/20 \\
Numeric     & 12/20 (L2)  & 0/20 & 10/20 (L4) & 0/20 \\
\bottomrule
\end{tabular}
\end{table}

\begin{figure}[t]
\centering
\includegraphics[width=\columnwidth]{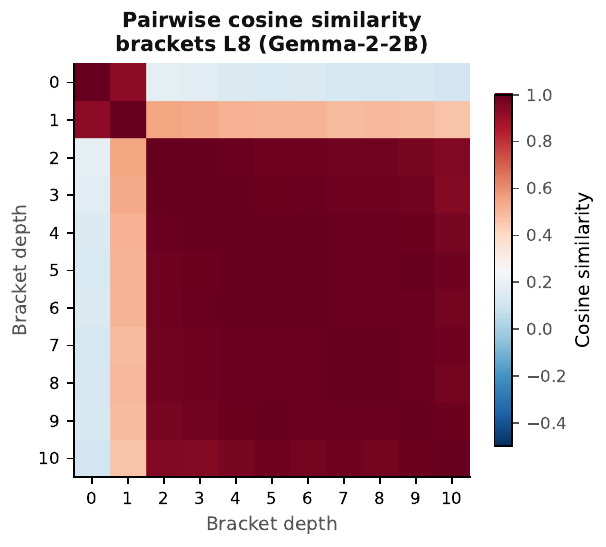}
\caption{\textbf{Pairwise cosine similarity of bracket depth mean activations} (Gemma-2-2B, L8). Depth 0 occupies a distinct direction; depths 2--10 are nearly identical, confirming a 1D manifold with one outlier at depth 0.}
\label{fig:cossim}
\end{figure}

\subsection{Geometric Computation and Subspace Ablation}

We analyzed all attention heads per model per task (\Cref{tab:twist}). The degree of geometric computation varies substantially across architectures, even when probing accuracy is similar.

\begin{table}[h]
\caption{Geometric computation summary. Max twist\_q score and residual alignment range (min$\to$max across layers). G-2-2B/9B = Gemma-2-2B/9B.}
\label{tab:twist}
\centering
\small
\begin{tabular}{@{}llcc@{}}
\toprule
Task & Model & Max twist & Resid.\ align. \\
\midrule
\multirow{3}{*}{Brackets}
 & G-2-2B & 0.033 & 0.78$\to$0.98 \\
 & G-2-9B & 0.060 & 0.72$\to$0.98 \\
 & Qwen3-4B & 0.131 & 0.65$\to$1.00 \\
\midrule
\multirow{3}{*}{Indentation}
 & G-2-2B & 0.214 & 0.47$\to$0.64 \\
 & G-2-9B & 0.331 & 0.43$\to$0.52 \\
 & Qwen3-4B & 0.803 & 0.46$\to$0.98 \\
\midrule
\multirow{3}{*}{Table rows}
 & G-2-2B & 0.215 & 0.44$\to$0.74 \\
 & G-2-9B & 0.135 & 0.62$\to$0.87 \\
 & Qwen3-4B & 0.237 & 0.51$\to$1.00 \\
\midrule
\multirow{3}{*}{Numeric}
 & G-2-2B & 0.210 & 0.32$\to$0.50 \\
 & G-2-9B & 0.288 & 0.32$\to$0.52 \\
 & Qwen3-4B & 0.724 & 0.37$\to$0.95 \\
\bottomrule
\end{tabular}
\end{table}

\begin{figure*}[t]
\centering
\includegraphics[width=\textwidth]{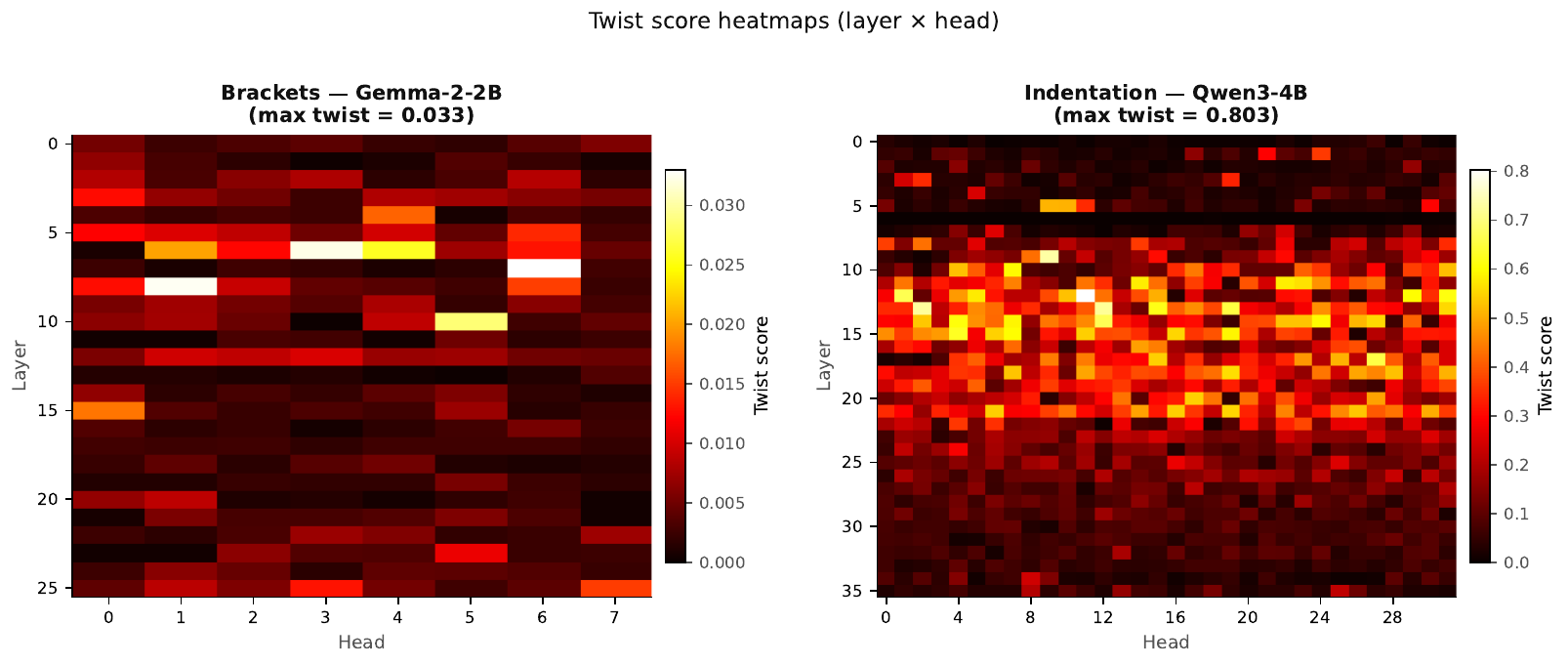}
\caption{\textbf{Twist score heatmaps (layer $\times$ head).} \emph{Left}: Brackets in Gemma-2-2B (max twist = 0.033) -- uniformly low, confirming pre-organized manifold. \emph{Right}: Indentation in Qwen3-4B (max twist = 0.803) -- concentrated in layers 5--20, with multiple ordinal-preserving twists. Note the 25$\times$ difference in scale.}
\label{fig:twist_heatmaps}
\end{figure*}

\paragraph{Pre-organized vs.\ constructed manifolds.} Bracket depth manifolds are pre-organized in all models: max twist $\leq 0.13$, residual alignment starting at 0.65--0.78 and climbing to near-perfect (\Cref{fig:twist_heatmaps}, left). We attribute this to bracket depth being directly encoded in token identity; the embedding layer already places tokens at the correct manifold position, so heads need only \emph{read} the manifold, not construct it. This contrasts with \citet{gurnee2026manifolds}'s character counting, where the count depends on all preceding characters. For indentation and table rows, residual alignment starts low and climbs through the network, indicating active manifold construction by attention heads.

\paragraph{Coherent vs.\ incoherent twisting.} The most notable cross-architecture difference is in twisting magnitude. Qwen3-4B shows 4--10$\times$ higher maximum twist than Gemma for indentation (0.803 vs.\ 0.214--0.331) and numeric magnitude (0.724 vs.\ 0.210--0.288). Crucially, Qwen3-4B's indentation twisters preserve ordinal order ($\rho \geq 0.77$), they rearrange the manifold while maintaining the ordinal progression, exactly the behavior \citet{gurnee2026manifolds} found for character counting. Its numeric twisters do not ($\rho \leq 0.23$), indicating incoherent rearrangement rather than structured geometric computation. The twist concentration ratio (max/mean) is similar across models ($\approx$5--8), meaning the difference is not specialization but rather that all Qwen3-4B heads twist more strongly in absolute terms, possibly reflecting its RoPE positional encoding or attention architecture (\Cref{fig:twist_heatmaps}, right).

\paragraph{Subspace ablation.} Activation patching confirms that the manifold subspaces are necessary for linear decodability (\Cref{tab:patching}). Patching the top-3 manifold directions causes large drops in probe accuracy, while random directions cause negligible drops ($<$0.01 except Qwen3-4B numeric). The ratio exceeds more than 28 times for every task--model combination. This is consistent with informational concentration in the identified subspace, not behavioral causality, whether patching shifts downstream generation remains an open question. Notably, even numeric magnitude, which lacks clean manifold structure, shows large manifold drops, indicating that high-dimensional representations can still concentrate task-relevant information in a small subspace. For Qwen3-4B numeric, the elevated random drop suggests this representation is less geometrically organized.

\begin{table}[t]
\caption{Activation patching results (top-3 manifold directions, best layer). Man.\ drop: decrease in test metric when patching manifold subspace. Rand.\ drop: decrease when patching random directions.}
\label{tab:patching}
\centering
\small
\begin{tabular}{@{}llccc@{}}
\toprule
Task & Model & Baseline & Man.\ drop & Rand.\ drop \\
\midrule
\multirow{3}{*}{Brackets}
 & Gemma-2-2B & 0.496 & 0.342 & 0.002 \\
 & Gemma-2-9B & 0.456 & 0.264 & $-$0.004 \\
 & Qwen3-4B   & 0.464 & 0.415 & 0.001 \\
\midrule
\multirow{3}{*}{Indentation}
 & Gemma-2-2B & 0.926 & 0.492 & $-$0.001 \\
 & Gemma-2-9B & 0.947 & 0.665 & 0.001 \\
 & Qwen3-4B   & 0.901 & 0.519 & $-$0.001 \\
\midrule
\multirow{3}{*}{Table rows}
 & Gemma-2-2B & 0.718 & 0.583 & 0.009 \\
 & Gemma-2-9B & 0.819 & 0.690 & 0.001 \\
 & Qwen3-4B   & 0.744 & 0.633 & 0.010 \\
\midrule
\multirow{3}{*}{Table cols}
 & Gemma-2-2B & 0.919 & 0.738 & $<$0.001 \\
 & Gemma-2-9B & 0.942 & 0.831 & $<$0.001 \\
 & Qwen3-4B   & 0.859 & 0.648 & 0.007 \\
\midrule
\multirow{3}{*}{Numeric}
 & Gemma-2-2B & 0.508 & 8.24 & 0.001 \\
 & Gemma-2-9B & $-$2.92$^\dagger$ & 20.0 & 0.025 \\
 & Qwen3-4B   & 0.243 & 22.2 & 3.10 \\
\bottomrule
\end{tabular}
\smallskip\\
{\footnotesize $^\dagger$ Gemma-2-9B numeric probe overfits severely (train $R^2=0.99$, test $R^2=-2.92$). The large manifold drop (20.0) reflects that patching disrupts the representation, but the absolute value should be interpreted cautiously given the broken baseline.}
\end{table}

\paragraph{Cross-task sharing.} Features \#9213, \#15567, \#7214, and \#6631 appear among the top-20 features for multiple tasks in Gemma-2-2B, but all are primarily value$=$0 detectors; better characterized as ``beginning of structure'' features than general ordinal position detectors. Cross-task overlap of top-20 twisting heads is modest (0--8 shared heads per pair) and model-dependent. No head appears in the top-20 twisters for all four tasks in any model.

%% ─── CONCLUSION ───
\section{Discussion and Conclusion}
\label{sec:discussion}

Our results suggest a tentative taxonomy of ordinal representations based on how the variable relates to token identity, though it is derived from only four tasks and awaits validation on additional domains. Tasks where the variable is \emph{locally computable} from token identity (bracket depth, table column index) produce clean 1D manifolds pre-organized from the embedding layer. Tasks requiring \emph{cross-position integration} (table rows, indentation) produce higher-dimensional manifolds (2--4D) that are actively constructed by attention heads. Tasks requiring \emph{semantic extraction} (numeric magnitude) produce diffuse, high-dimensional representations (4--5D) without clean manifold structure, despite strong linear decodability. This distinction between decodability and geometric structure is important: our ``manifold dimensionality'' refers to the centroid manifold (PCA on per-value means), and low centroid dimensionality does not guarantee that token-level representations concentrate near a low-dimensional curve, though the subspace ablation results indicate these directions are informationally privileged.

The degree of geometric computation is architecture-dependent: Qwen3-4B shows 4--10$\times$ stronger twisting than Gemma for tasks requiring manifold construction, with its indentation twisters preserving ordinal order while its numeric twisters do not. This may reflect RoPE positional encoding or architectural differences rather than head specialization, since the twist concentration ratio is similar across models.

\paragraph{Limitations.}
The bracket depth probe overfits substantially (99\% train, 53--57\% test). The synthetic bracket data has a minor labeling inconsistency (always closing with `)` regardless of opener). No pretrained SAEs exist for Qwen3-4B, so the two central findings, place-cell feature tiling and strong geometric twisting, are demonstrated on different models; training an SAE on Qwen3-4B to bridge this gap is an important next step. The twist score operates on per-value \emph{mean} activations and may miss transformations on individual tokens. Our subspace ablation establishes informational concentration, not behavioral causality.

\paragraph{Future directions.}
Training SAEs on Qwen3-4B would test whether stronger geometric computation corresponds to cleaner feature tiling. The information-geometric framework of \citet{park2026infogeo} may better characterize curved manifolds than Euclidean PCA. Extending to larger models (Gemma-2-27B, Qwen3-14B) would test whether architecture-dependent differences scale with model size. Behavioral causality experiments, testing whether patching subspaces shifts downstream generation, would strengthen the causal interpretation of our findings.

\bibliography{references}
\bibliographystyle{icml2026}

\end{document}